# Invariant properties of a locally salient dither pattern with a spatial-chromatic histogram

A. M. R. R. Bandara, L. Ranathunga, and N. A. Abdullah, *Member, IEEE*

Original Paper:

A. M. R. R. Bandara, L. Ranathunga and N. A. Abdullah, "Invariant properties of a locally salient dither pattern with a spatial-chromatic histogram," 2013 IEEE 8th International Conference on Industrial and Information Systems, Peradeniya, 2013, pp. 304-308.
doi: 10.1109/ICIInfS.2013.6732000
keywords: {image colour analysis;image retrieval;pattern classification;support vector machines;CDPC;SVM classifier;compacted dither pattern code;invariant properties;linear support vector machine;locally salient dither pattern;rotational invariance property;scale invariance property;spatial-chromatic histogram;visual depiction;Feature extraction;Histograms;Image color analysis;Information technology;Shape;Support vector machines;Visualization;Compacted Dither Pattern Codes;Content Based Retrieval;SVM;Salient Dither Pattern Feature},
URL: http://ieeexplore.ieee.org/stamp/stamp.jsp?tp=&arnumber=6732000&isnumber=6731935

# Invariant Properties of a Locally Salient Dither Pattern with a Spatial-Chromatic Histogram

A. M. R. R. Bandara, L. Ranathunga, and N. A. Abdullah, *Member, IEEE*

*Abstract*- Compacted Dither Pattern Code (CDPC) is a recently found feature which is successful in irregular shapes based visual depiction. Locally salient dither pattern feature is an attempt to expand the capability of CDPC for both regular and irregular shape based visual depiction. This paper presents an analysis of rotational and scale invariance property of locally salient dither pattern feature with a two dimensional spatial-chromatic histogram, which expands the applicability of the visual feature. Experiments were conducted to exhibit rotational and scale invariance of the feature. These experiments were conducted by combining linear Support Vector Machine (SVM) classifier to the new feature. The experimental results revealed that the locally salient dither pattern feature with the spatial-chromatic histogram is rotationally and scale invariant.

*Index Terms*— Compacted Dither Pattern Codes, Content Based Retrieval, Salient Dither Pattern Feature, SVM

## I. Introduction

Due to the rapid increase of the amount of multimedia contents in the internet, multimedia productions, surveillance and medical applications, "Multimedia Content Search and Retrieval" has become an active research field [1-4]. The text based visual retrieval is one of the main approaches to retrieve visual data. In this approach, a textual representation of the query image or video is obtained using a visual feature descriptor. Color, texture and shape are the commonly used visual features and a single or a collection of these features is used to create a visual descriptor[2].

Color feature is invariant to image size and orientation. Color histograms[5] and color correlogram[6] are the mostly used color descriptors. The color correlogram addresses the issue of lacking spatial information in color histograms. However, correlograms tend to have high dimensionality due to the requirement of storing pair of colors with respect to their neighbor distances[6]. Texture features give information about the spatial arrangement of colors or intensities in an image or selected region of an image[7]. Homogeneous Texture Descriptor (HTD)[8], Texture Browsing Descriptor (TBD)[9] and Edge Histogram Descriptor (EHD)[10] are some of conventional texture descriptors. These descriptors have their own context of applications. This fact yields that the texture descriptors are strictly application dependant. Unlike the color and texture features, the shape feature can directly describe objects in visual data. There are generally two types of shape descriptors: contour-based shape descriptors and region-based shape descriptors. Contour-based shape descriptors exploit only boundary information and cannot deal with disjoint shapes where boundary may not be available. Therefore, they have limited applications. In region based techniques, all the pixels within a shape region are taken into account to obtain the shape representation. Common region based methods use moment descriptors to describe shape such as geometric moments, Legendre moments and Zernike moments[11]. Shape information retrieval was limited to specialized retrieval systems.

Besides the fundamental visual features there are extended local features that are formed with combinations of the fundamental features. Within the past decade, researchers have invented several significant local features and descriptors such as Scale Invariant Feature Transform(SIFT)[12] and Speeded Up Robust Feature (SURF)[13]. SIFT and SURF are the most popular local feature descriptors but both are suffered from high dimensionality and computational complexity[14]. Compact Dither Pattern Codes (CDPC) is another successful feature descriptor which is found in recent years. CDPC is applicable to irregular shapes based video visual concept depiction which includes both color and texture information in compacted manner[15].

CDPC has addressed many of the above mentioned problems namely high dimensionality of multi-feature descriptors and high computational complexity of visual feature extraction. However, CDPC does not include shape information. We attempt to extend the capability of CDPC concept by incorporating shape information through a locally salient dither pattern feature with a spatial-chromatic histogram technique. A visual feature descriptor should have many invariant properties such as rotational invariance, scales invariance and illumination invariance etc. in order to be robust in applications. In this paper, we describe an analysis of rotational and scale invariance property of the proposed technique which expands the applicability of the proposed visual feature over a wide range of applications. The previous studies that are related to the locally salient dither pattern feature and the formation of spatial-chromatic histogram will be explained in the next section.

A. M. R. R. Bandara is with the Department of Information Technology, Faculty of Information Technology, University of Moratuwa, Sri Lanka (e-mail: ravimalb@uom.lk).
  L. Ranathunga is with the department of Information Technology, Faculty of Information Technology, University of Moratuwa, Sri Lanka (e-mail: lochandaka@uom.lk).
  N. A. Abdullah is with Department of Computer System and Technology, Faculty of Computer Science & Information Technology, University of Malaya, 50603 Kuala Lumpur, MALAYSIA. (e-mail: noraniza@um.edu.my)

## II. RELATED STUDIES

### A. Compact Dither Pattern Codes (CDPC)

CDPC is one of the syntactic visual features for irregular shapes based video visual concept depiction. It focused on reducing the chromatic depth and spatial density by applying a dithering technique. In CDPC, four codes are generated for each four color blocks which are adjacent to each other. It contains some properties of texels as it represents combinational chromatic information over a tiny square space. The CDPC grid size of $l=2x2$ square dither color points, in which each of dither point in CDPC grid represents, $t^2$ pixels of block[15]. If the reduced depth of colors is $z$ then the number of different color dither pattern grids created will be given by (1),

$$N = \binom{z}{4} + \binom{z}{2} + \sum_{r=0}^{2}\binom{z}{r}\cdot(z-r) \quad (1)$$

Where $x>=4$. The $N$ represents the selection of the number of elements in CDPC set.

Due to the spatial blending effect in dither patterns, CDPC has been able to successfully reduce the chromatic depth without affecting the overall visual impression. The each extracted four codes sets are arranged in descending order in order to exclude internal permutations within the pattern. This Process reduces the number of different color dither pattern grids. As the final output a descriptor is generated with the probability distribution of the CDPC patterns in a given spatial area hence it includes only color and texel properties which can depict irregular shape based concepts[15].

The resultant descriptor is used for the semantic retrieving process. CDPC shows improvements in retrieval accuracy, dimensionality of the descriptor and computational complexity in irregular shape based video visual detection and description. CDPC has been compared with MPEG-7 Dominant Color Descriptor (DCD) and it shows better results[15].

The new feature description includes the spatial information using the centroid distance function, which will be explained in the next section.

### B. Centroid Distance Function

Centroid distance function is a one method that comes under one-dimensional and non-reconstructive shape representation. it has been used for shape representation in several studies[16,17]. The centroid or the centre of gravity is calculated based on the way that the shape is represented. If the shape is represented with region function, the centroid $(x_c, y_c)$ is given by (2),

$$x_c = \frac{1}{N}\sum_{i=1}^{N} x_i$$
$$y_c = \frac{1}{N}\sum_{i=1}^{N} y_i \quad (2)$$

Where $N$ is the number of points in the shape, $(x_i, y_i) \in \{(x_i, y_i) | f(x_i, y_i) = 1\}$.

The centroid distance function is expressed by the distance of the boundary points $(x(t), y(t))$ from the centroid $(x_c, y_c)$ of the shape as in (3).

$$r(t) = ([x(t) - x_c]^2 + [y(t) - y_c]^2)^{\frac{1}{2}} \quad (3)$$

Due to the subtraction of centroid, which represents the position of the shape, from boundary coordinates, centroid distance representation is invariant to translation. It also resists to noises and occultation. Centroid distance function representation also has very low computational complexity[18]. We use centroid distance function to obtain the descriptor which will be explained in next section.

## III. PROPOSED FEATURE AND DESCRIPTOR

### A. Locally Salient Dither Pattern Feature Extraction

We take the advantage of the fundamental idea of CDPC, which the dithering reduces both color and spatial data density while preserving the visual impression. Further we consider only about locally salient dither patterns instead of the ones with highest probability over the specified spatial region. This adoption yields a point like feature as well as suppresses redundant patterns over flat regions.

The dither pattern feature extraction involves several sub processes which are explained under four distinguished steps explained below.

*Step 1*: Apply neighborhood color averaging filter for each $t^2$ pixels of blocks. This step outputs an image with lower resolution and less noise.

*Step 2*: A dither pattern is a square shape region formed by four blocks, which's size is $2t \times 2t$. Each pattern contains four color values.

Fig. 1 shows the structure of dither pattern with the four blocks. Any of four adjacent blocks in the image is considered as a dither pattern. The pattern which has a great dissimilarity over its neighbor patterns is defined as a salient dither pattern. We use the summation of absolute color differences between the pattern and its neighbor patterns in order to detect salient dither patterns which have the potential to be the feature points.

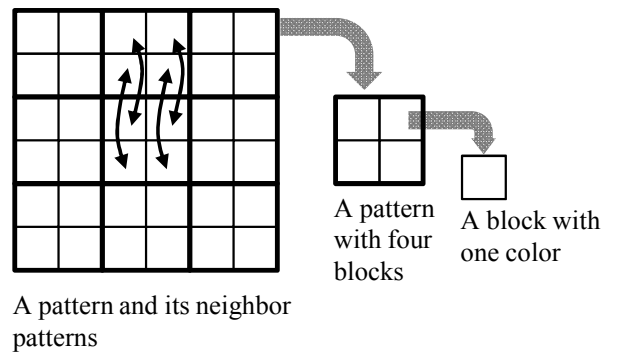

Fig. 1. Structure of the dither pattern

Let $D_p$ be the color distance from a $p^{th}$ neighbor pattern ($p = 1,2,...,8$) to the current pattern $n$ and is calculated using (4).

$$D_p = \sum_{i=1}^{4}\sum_{c=1}^{3}|C_c(E_i(P_p)) - C_c(E_i(P_n))| \quad (4)$$

Where $C_c(E_i(P_p))$ is the $c^{th}$ color component of the $i^{th}$ element in the $p^{th}$ neighbor pattern. The function $C_c$ is in Red-Green-Blue (RGB) space hence red, green and blue components are corresponded to the 3 values of $c$.

Then we check whether the pattern is dissimilar to all its neighbor patterns. We determine whether a pattern is a potential feature point or not, by applying a threshold to the distances from all its neighbor patterns. We select the threshold value empirically where it does not detect edges but corners. After applying the threshold, the selected patterns are kept in a set $S_s$, the set of patterns which are potential feature points.

*Step 3*: The vector $\{D_1, D_2...\}$ signifies the saliency of a pattern over its neighbor patterns. This saliency expresses the likelihood of a selected pattern to be a feature. This likelihood is called as feature strength. We express the overall strength of a feature point as a summation of the $D_p$ values.

*Step 4*: We observed that the points in $S_s$ are concentrated to several small regions in an image. Therefore we need to reduce the number of feature points while keeping the higher strength feature points in each of the regions. In order to doing that we suppress the points with non-maximal feature strength from the interested regions to select only the best feature points. A fix sized window is considered to collect feature points for finding the maximum strength. After suppressing the non-maximal feature points, the remaining points are kept in a set $S_f$, the set of salient dither pattern feature points.

We integrate the spatial details and chrominance details of the set of feature points into a 2D histogram to build the descriptor. One axis of the histogram shows the spatial information whereas the other axis shows the color information. This 2D histogram contains not only the spatial and color information but also the correlation of both spatial and color information of the set of feature points.

### B. Spatial Details Extraction

We use centroid distance function since it reduces the spatial representation from 2D to 1D and its invariant properties. Centroid distance function is not scale invariant[18]. We adopt the function in order to make it scale invariant and integrate the spatial details of the set of feature points to a histogram. We calculate the centroid for the set of feature points in $S_f$ using (2) since the distribution of feature points over spatial region is more likely a region based shape representation.

After calculating the centroid, the distance to each feature point from the centroid is calculated. Based on the distances, the feature points are classified into one of the distance bins in the histogram. We allocate the bins for centroid distances of the feature points as the distance range of each bin covers the same amount of area.

Let $D_f(i)$ is the spatial distance from $i^{th}$ feature point to the centroid and $\max(D_f^2(i))$ denotes the maximum of squared $D_f(i)$ values; $k$ is the number of bins that we are going to prepare. We define the bin range for the $n^{th}$ bin ($Bin(n)$) as in (5). We build the function $Bin(n)$ as it assures that each bin corresponds to the same amount of area in the image.

$$Bin(n) = \begin{cases} \dfrac{\max(D_f^2(i))}{k}, & n = 1 \\ Bin(n-1) + \dfrac{\max(D_f^2(i))}{k}, & 1 < n \leq k \end{cases} \quad (5)$$

The conventional way to calculate the centroid distance function $r(i)$ is given in (3). But the histogram requires the squared of centroid distance ($D_f^2(i)$) and it is equal to the squared of function $r(i)$. This adoption omits a large number of square root operations hence it saves computational time. After preparing the bins, each feature point in $S_f$ is allocated to a one of the $k$ number of bins based on its centoid distance. Then we reduce the color depth of dither patterns of feature points in each distance bin.

### C. Color Reduction

The color quantization is done in Hue-Saturation-Value (HSV) color space in order to reduce the number of bins in the histogram. We quantize the color into fewer levels as shown in Fig. 2 and assign to the color bins. This quantization method results in the color histogram to capable of handling not only the color images but also grey images.

### D. Spatial-Chromatic Histogram

We create the color histograms for each distance bin, using the frequency of colors in dither patterns of the allocated feature points. We concatenate these histograms to form a 2D histogram which contains one axis for the color bins and another axis for the distance bins.

We rescale the histogram to the range of 0 to 1 by dividing the each element by the maximum value of the histogram. The resultant histogram is used for semantic classification of images with the help of support vector machine (SVM).

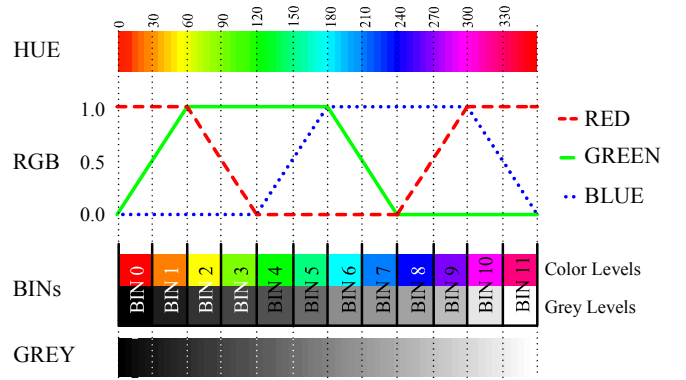

Fig. 2. The color quantization and mapping to the color bins.

## IV. EXPERIMENTAL RESULTS

We setup two experiments to test the rotational invariance and scale invariance properties of the salient dither pattern feature with 2D spatial-chromatic histogram. We use Corel dataset[19] for the experiments. Corel dataset includes 1000 images which are labeled into ten categories as each category contains 100 images. Fig 3 shows some images of Corel dataset. We checked the retrieval rate of images from Corel dataset with the all combination of number of distance bins $k$ (range 3-10) and number of color levels (values: 6, 12 and 24). It showed that the maximum retrieval rate achieved at the point (4, 12) hence we set $k$ to 4 and number of color levels to 12 for our experiments.

### A. Rotational Invariance

In order to test the rotational invariance we created 360 images by rotating a single source image over 360 degrees each with a single degree difference. A few examples for the rotated images are shown in Fig. 4 Then their feature vectors are extracted. A two class linear SVM is trained with 6 randomly selected images from the set of 360 images as for the first class and 99 other similar images for the other class. For an example if an image of a bus is selected as the source image then the similar images will also contain buses as shown in the Fig. 3 We obtain the confidence value (the probability estimate) from the SVM by predicting the class for all 360 images. If an image is matched to a class then the confidence value of matching the image for the same class will be greater than 0.5. This process was carried for two source images from two different image categories. Fig. 6 and Fig. 7 show the confidence value of matching the rotated images of the two source images. Then we calculate the average of the confidence values obtained using the two source images. The graph in Fig. 8 shows the average confidence values obtained for the two source images.

### B. Scale Invariance

In order to test the scale invariance we created 8 images by scaling a single image with the scaling factors 0.25, 0.5, 0.75, 1, 1.25, 1.5, 1.75 and 2. A few examples of scaled images are shown in Fig. 5. A two class linear Support Vector Machine (SVM) is trained with 1 randomly selected image from the set of 8 images as for the first class and 99 other similar images for the other class. We obtain the confidence value by predicting the class for all 8 images. This process was carried for ten source images from ten different image categories and calculated the average of the confidence values for the ten source images. The graph in Fig. 9 shows the average of confidence values of matching the source images to the given class against scaling over the eight scale factors.

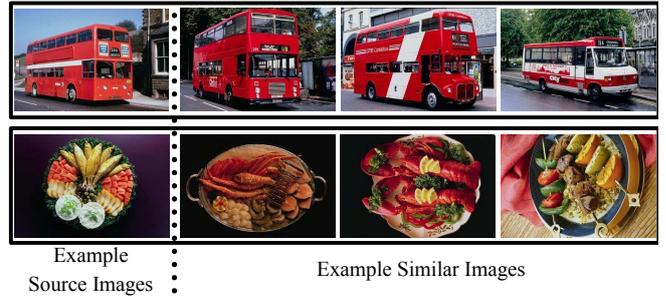

Example Source Images — Example Similar Images

Fig. 3. Example images of the database

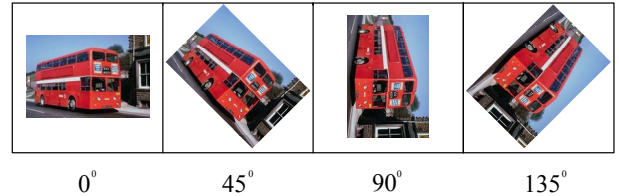

0°   45°   90°   135°

Fig. 4. Example rotated images with the degrees of rotation

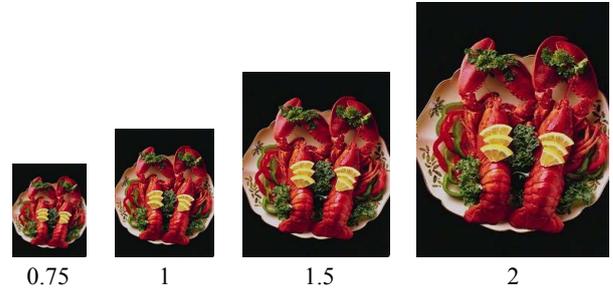

0.75   1   1.5   2

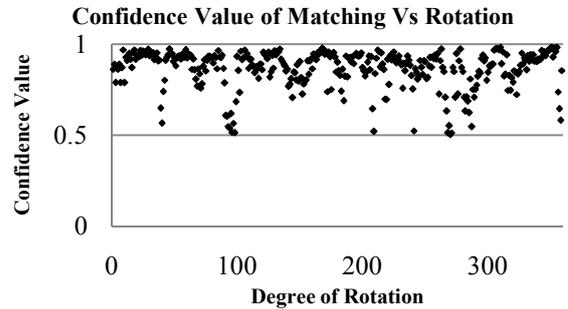

Fig. 6. Confident values of matching with rotated images to its class – for the rotated images of the first source image

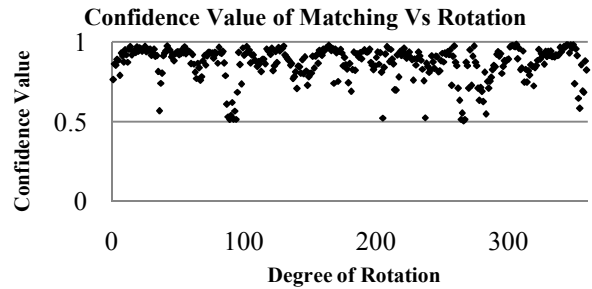

Fig. 7. Confident values of matching with rotated images to its class – for the rotated images of the second source image

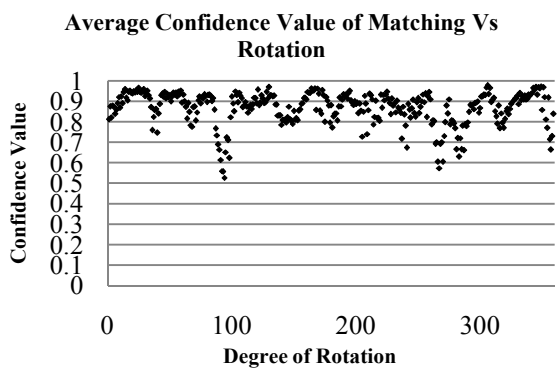

Fig. 8. Average confidence values of matching rotated images to its class

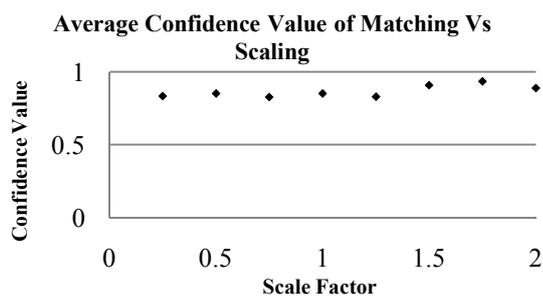

Fig. 9. Average confidence values of matching scaled images to its class

## V. DISCUSSION AND CONCLUSION

The confidence value of any given matching of rotated image for its class is greater than 0.5, hence 100% of the rotated images are matched to their class. The average of the confidence values of matching for all rotated images is 0.87 which shows a great likelihood of the feature description of rotated images to its class which is trained using only 2% of the set of 360 images and 99 other similar images. The confidence values of matching are much lower than the average for some degrees of rotation. The graphs in Fig. 6 and Fig. 7 show similar fluctuations of confidence value whereas the degrees near 100, 300 and 360 show significantly lower confidence values. Further studies are required to identify the causes for this nature of fluctuation and to improve the confidence value.

The confidence value of any given matching of scaled image for its class is greater than 0.5 which clearly shows that 100% of the scaled images are matched to their class. The average of the confidence values of matching for all 8 images is 0.87 which shows a great likelihood of the feature description of scaled images to its class which is trained using only one of the set of 8 images and 99 other similar images.

In this paper, we described the analysis of rotational and scale invariance property of a locally salient dither pattern feature with a 2D spatial-chromatic histogram. We used the centroid distance function in order to maintain the rotational invariance property of the histogram. We also introduced the scale invariance property to the histogram by scaling it to a fixed range. The study revealed that locally salient dither pattern feature with the spatial-chromatic histogram descriptor is invariant to rotation and scale transformation with 0.87 of average confidence value of matching. This property expands the applicability of the proposed visual feature over a wide range of applications. Further work of this research is to evaluate the visual feature with the descriptor against state-of-the-art visual feature descriptors with various datasets.

ACKNOWLEDGMENT

This work was carried out with the support of National Research Council, Sri Lanka. (Grant No. 12-017)